\def\year{2021}\relax
\documentclass[letterpaper]{article} 
\usepackage{aaai20}  
\usepackage{times}  
\usepackage{helvet} 
\usepackage{courier}  
\usepackage[hyphens]{url}  
\usepackage{graphicx} 
\usepackage{multirow}
\usepackage{graphicx}  
\usepackage{amsmath,amssymb,amsthm,bbm,mathtools,bm}
\usepackage{algorithm,algorithmicx,algpseudocode}
\usepackage{subfigure}
\usepackage{color}

\graphicspath{{../docs/figs/}}

\title{Deep Multi-View Spatiotemporal Virtual Graph Neural Network for Significant Citywide Ride-hailing Demand Prediction}
\author{
Guangyin Jin,$^{1}$\thanks{Contact author}
Zhexu Xi,$^{2}$
Hengyu Sha,$^{1}$
Yanghe Feng,$^{1}$
Jincai Huang,$^{1}$\footnotemark[1]\\
$^1$College of Systems Engineering, National University of Defense Technology\\
$^2$Bristol Centre for Functional Nanomaterials, University of Bristol\\
jinguangyin18@nudt.edu.cn, hu20363@bristol.ac.uk, \{shahengyu,fengyanghe,huangjincai\}@nudt.edu.cn
}

\begin{document}
\maketitle
\begin{abstract}
Urban ride-hailing demand prediction is a crucial but challenging task for intelligent transportation system construction. Predictable ride-hailing demand can facilitate more reasonable vehicle scheduling and online car-hailing platform dispatch. Conventional deep learning methods with no external structured data can be accomplished via hybrid models of CNNs and RNNs by meshing plentiful pixel-level labeled data, but spatial data sparsity and limited learning capabilities on long-term temporal dependencies are still two striking bottlenecks. To address these limitations, we propose a new virtual graph modeling method to focus on significant demand regions and a novel Deep Multi-View Spatiotemporal Virtual Graph Neural Network (DMVST-VGNN) to strengthen learning capabilities of spatial dynamics and temporal long-term dependencies. Specifically, DMVST-VGNN integrates the structures of 1D Convolutional Neural Network, Multi Graph Attention Neural Network and Transformer layer, which correspond to short-term temporal dynamics view, spatial dynamics view and long-term temporal dynamics view respectively. In this paper, experiments are conducted on two large-scale New York City datasets in fine-grained prediction scenes. And the experimental results demonstrate the effectiveness and superiority of DMVST-VGNN framework in significant citywide ride-hailing demand prediction.
\end{abstract}

\section{Introduction}
With the development of unbanization, messy traffic management has been deteriorating the daily life, like serious traffic congestion. Therefore, we expect an intelligent transportation system (ITS) that can give precise demand predictions on travel by providing intelligent scheduling decisions. Urban ride-hailing demand prediction is a crucial part in ITS, which is a useful tool for data-driven traffic management.

In spatiotemporal deep learning models, there are two most common types of spatial data representation, image-based modeling approach and graph-based one. The image-based approach is to grid urban areas evenly referring to latitude and longitude, where the statistical data in each grid can be regarded as the pixel value of the pixel-level image. In this way, the image-based spatial data can be recognized by CNN-based models \cite{xingjian2015convolutional,yao2018deep,yao2018modeling,zhang2017deep}.However, the biggest disadvantage of this means is the difficulty in controlling the gird granularity. High-granularity can more accurately capture the dynamics of fine-grained regions but triggers the problem of spatial data sparsity, which makes it difficult to learn and optimize parameters in CNN-based models. In contrast, low-granularity can reduce the sparsity of spatial distribution but could lead to mergence of many heterogeneous regions, eliminating practical significance in prediction. The graph-based approach was mainly utilized in the defined graph scenes, for instance, road networks and sensor networks, whose spatial dynamics can be grasped by GCN-based models \cite{zhao2018temporal,zheng2020gman,li2017diffusion,yu2017spatio,zhang2020deep}. There are still some points needing mentioning: the lack of access to structured data, and low transfer rate by various scene-based structured information. Thus, we only use the ride-hailing GPS data without any external structured data in this paper. Also, we grid urban areas evenly with finer granularity according to latitude and longitude, then discard the sparse signal regions with a certain threshold, and maintain the significant ones. This is because significant demand signals show the pattern of urban ride-hailing while the insignificant ones often represent  abnormal signals in some special terrain(mountains, wilderness, lakes, etc.). To enhance the practical significance of prediction, we propose a spatiotemporal aggregation method where we turn a collection of retained neighbor regions with high ride-hailing pattern similarity into new virtual nodes. In addition, we adopt three different views, distance, correlation and mobility graph, to model the virtual graph. 

In most previous research, RNN and its variants are fruitful models to capture temporal dynamics from sequential data but they still have limited processing capability for long-term dependencies \cite{xingjian2015convolutional,yao2018deep,cui2019traffic,yao2018modeling,jin2019crime,wang2020csan}. In this paper, we propose a Deep Multi-View Spatiotemporal Virtual Graph Neural Network (DMVST-VGNN), which is stacked by three hierarchical structures-- short-term temporal dynamics view, spatial dynamics view and long-term temporal dynamics view. Gated 1D convolution operation is involved in the first hierarchical structure, short-term temporal dynamics view, which has comparable effect and lower computing cost compared with RNN \cite{dauphin2017language}. Then the multi-type spatial dependencies are extracted in spatial dynamics view by multi Graph Attention Neural Network (GAT). Followed by spatial dynamics view, Transformer \cite{vaswani2017attention} model is applied to grasp long-term dependencies in long-term temporal dynamics view. In this framework, the coupling association of short-term dependencies, spatial dependencies and long-term dependencies is established.In summary, our contributions are summarized as follow:
\begin{itemize}
\item It is the first exploration to model virtual graphs of significant citywide ride-hailing demand regions without any external data, to the best of our knowledge. Meanwhile, our graph modeling approach is more interpretable and overcomes the difficulties of spatial sparse data distribution in demand prediction.
\item We put forward the novel multi-view deep learning model that combines short-term temporal dynamics view, spatial dynamics view and long-term temporal dynamics view together.
\item We evaluate our model with two real-world datasets. Superior performance on significant citywide ride-hailing demand prediction of our proposed model have been fully demonstrated.
\end{itemize}

\section{Related Work}
Early research on traffic forecasting mainly focuses on mathematical modeling methods and statistical learning methods. One of the mainstream mathematical-modeling methods is classic time series modeling, Autoregressive Integrated Moving Average (ARIMA) and its variants have been widely applied in traffic prediction tasks \cite{han2004real}. To strengthen the model's generalization performance, some ensemble statistical learning methods were proposed, for instance, Random Forest, Adaboost and Xgboost. These theories have all been successfully used in traffic state prediction \cite{leshem2007traffic,leshem2007traffic}. Although statistical learning and mathematical modeling methods are interpretable, it is hard for them to capture complex dynamics from spatiotemporal scale.

More recently, deep learning models have been increasingly more widely employed in traffic state prediction. ConvLSTM model \cite{xingjian2015convolutional} was a pioneering work in this field. Based on this work, more and more improved CNN and RNN coupling structures are applied in spatiotemporal prediction tasks. Initially, some urban areas, road networks and sensor networks were divided into form of pixel images as the inputs of CNN structures \cite{yao2018modeling,liu2017short}. This series of methods are convenient but they ignore internal structured information in traffic networks. Accordingly, ConvLSTM drove the proposal of GCRNN, where graph convolution operation replaced the original convolution operation \cite{seo2018structured}. This coupling structure can capture spatiotemporal dynamics on traffic network and has developed some effective variants, such as TGCN, DCRNN \cite{li2017diffusion,zhao2018temporal}. However, as discussed above, RNN structures have limited capability in learning long-term dependencies as well as difficulties in optimization. To overcome these limitations, some alternative methods have been explored in recent two years, for instances, Gate 1D CNN,spatiotemporal attention mechanism and spatiotemporal synchronization graph \cite{wu2019graph,song2020spatial-temporal,zheng2020gman}. We also abandoned RNN structures in this paper, but the difference is that we divided the temporal learning unit into two parts: short-term temporal dynamics view and long-term temporal dynamics view. Motivated by the most recent work \cite{wang2020traffic}, we introduced Transformer structure to capture long-term dependencies in our work but replaced RNN structures with Gate 1D CNN as short-term dependencies learner, which is easier for optimization.

From the previous literatures, we can find that traffic flow prediction and traffic speed prediction focus more on predicting the traffic state via urban backbone road network or sensor network but traffic demand prediction places emphasis on citywide segmentation. Therefore, the spatial representation of ride-hailing demand is commonly image-based. However, in this paper, we extracted significant demand regions to construct three virtual graphs: distance graph, mobility graph and correlation graph, and solved the spatial sparsity problem in fine-grained prediction. The idea of multi-graph modeling we mentioned was also proposed in the works \cite{chai2018bike,geng2019spatiotemporal,jin2020urban,zhang2020multi}, yet on the basis of pre-defined stations, POI or any external structured information. Most importantly, without any support, we are able to achieve lower data dependency with the construction of GPS graph-based data in our work.

\section{Preliminaries}
\vspace{-0.05cm}
\noindent\textbf{Definition1.} Ride-hailing demand prediction: the task is to predict the next time slot citywide ride-hailing demand $Y_{t+T}$ from historical demand data $[Y_{t+1},Y_{t+2},¡­,Y_{t+T-1}]$.

\noindent\textbf{Definition2.} Spatiotemporal graph: Given a set of graphs $G_t=(g,x_t)$. Where $g=(v,e,w)$ represents a graph with vertex set \emph{v}, edge set \emph{e} and adjacency matrix \emph{w}. The notation $x_t$ represents the graph signals/features of each vertex. In this paper, the structure of \emph{g} is static, which represents the spatial correlations between different vertices. But $x_t$ is dynamics, which represents the dynamics feature of demand in each vertex that changes over time.

\section{Methodology}
The proposed deep learning framework DMVST-VGNN is displayed in Figure 1. It mainly consists of four components: 1) Graph generation, which aims to construct spatiotemporal virtual graph from distance, correlation and mobility view. 2) Short-term temporal dynamics view, which aims to capture short-term temporal dependencies by Gate 1D CNN for each vertex. 3) Spatial dynamics view, which aim to capture the spatial relations by Stacked GAT layers between different vertices; and 4) Long-term temporal dynamics view, which aims to capture long-term temporal dependencies by transformer layer for each vertex.

Note that, we use a feedforward neural network (FNN) to map the original input to a higher-dimensional space to enhance the representation. And there is a residual connection between the high-dimensional representation and output from spatial dynamics view, which merges the shallow and deep information of the model. In the following sub-sections, we will introduce the details of each component.
\begin{figure*}[htpb!]
  \centering
  \includegraphics[width=0.9\linewidth]{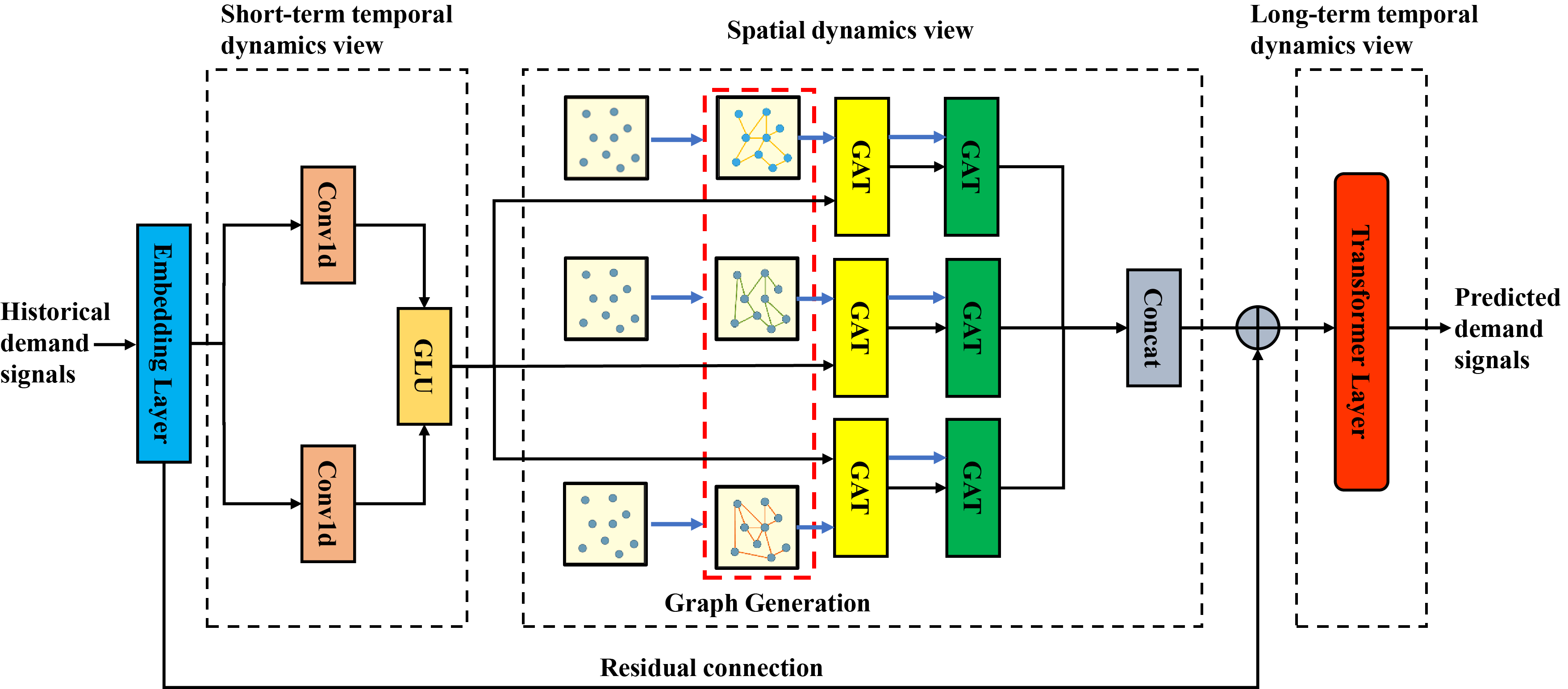}
  \setlength{\abovecaptionskip}{0pt}
  \setlength{\belowcaptionskip}{0pt}
  \caption{The overview of our proposed framework DMVST-VGNN. The black arrow lines represent the inflow of the graph signals and the blue arrow lines represent the inflow of the graph adjacency matrices.}
\end{figure*}

\subsection{Graph generation}
Most previous citywide spatiotemporal prediction task is to equally grid the urban area into small segmentation according to longitude and latitude. As we discuss above, this approach could lead to many heterogeneous regions or sparse regions, which reduces the accuracy and interpretability of prediction. Hence, we propose a novel virtual graph generation method to address the problem, which convert the image-based spatial representation to graph-based representation. The virtual graph generation method is divided into three steps, discarding regions, aggregating regions, constructing graphs respectively.

The first step, discarding regions, is to discard some regions with extremely sparse demands at a certain threshold. In citywide spatiotemporal prediction, what we are more concerned about is to capture the significant spatiotemporal dynamics rather than being disturbed by some occasional events. In high-granularity grid scene, many regions are statistically sparse, which involve noises and optimization problems for deep learning models. Given citywide segmentation divided by longitude and latitude $S=[s_1,s_2,\cdots ,s_n]$, we can calculate the average value of the demand time series $t\in[1,T]$ for each region i in the training set:$\overline{d_i}=\frac{1}{t}(d_1^i+d_2^i+\cdots+d_T^i)$, then discard the sparse demand regions by $\overline{d_i}<\delta$ and get a new region segmentation set $S_{new}=[s_1,s_2,\cdots,s_m ] (m<n)$.

The second step, aggregating regions, is to aggregate some regions with similar spatiotemporal patterns. Considering the interpretability of the prediction, our aim is to finely distinguish regions with different patterns, avoiding the heterogeneous regions. But fine-grained grids could also separate some neighbor regions with similar patterns. To strengthen practicality, we aggregate some similar neighbor regions in this step. The aggregation operation is based on the Pearson similarity of different regions, and the specific details are shown in Table 1.

The last step, constructing virtual graphs, is to construct distance graph, correlation graph and mobility graph based on the screened and aggregated regions. The reciprocal of distances, Pearson coefficients and mobility between different aggregated regions are computed in this step. For each region, we select the top 10$\%$ regions to connect to construct the adjacency matrices of distance graph, correlation graph and mobility graph.

\noindent We can construct the adjacency matrix of correlation graph as:
\begin{equation}A_{ij}^c=\left\{\begin{array}{ll}
1 & \text { If $r_{ij}$  is the top $10\%$} \\
0 & \text { otherwise }
\end{array}\right.\end{equation}
We can constructing the adjacency matrix of distance graph as:
\begin{equation} A_{ij}^d=\left\{\begin{array}{ll}
1 & \text { If $\frac{1}{d_{ij}}$  is the top $10\%$} \\
0 & \text { otherwise }
\end{array}\right.\end{equation}
We can construct the adjacency matrix of mobility graph as:
\begin{equation} A_{ij}^m=\left\{\begin{array}{ll}
1 & \text { If $m_{ij}$  is the top $10\%$} \\
0 & \text { otherwise }
\end{array}\right.\end{equation}

\begin{table}[htpb!]
\setlength{\abovecaptionskip}{-10.pt}
\caption{Process of spatiotemporal aggregation method}
\begin{tabular}{|l|}
\hline
 Spatiotemporal aggregation method\\ \hline
 1.	Calculating temporal similarity by Pearson coefficients\\ between different regions¡¯ demand time series in  $S_{new}$\\
 \qquad $r_{ij}=\frac{\sum_{t=1}^T(d_t^i-\bar{d_i})(d_t^j-\bar{d_j})}{\sqrt{\sum_{t=1}^T(d_t^i-\bar{d_i})^2}\sqrt{\sum_{t=1}^T(d_t^j-\bar{d_j})^2}}$\\
 2. Constructing segmentation label set\\ \qquad$S_{label}=[l_1,l_2,\cdots,l_m ]$ $(l_i=0,i=1,\cdots,m)$\\
 3. \textbf{For} each segmentation $s_i$ in $S_{new}$:\\
 4. \quad \textbf{If} $l_i\neq 0$: continue \\
 5. \quad \textbf{Else:} Finding its spatiotemporal neighbor set that \\ \qquad\quad satisfy both spatial adjacency and similar temporal \\ \qquad\quad patterns $(r>\epsilon):N_i$\\
 6. \quad \textbf{If} $N_i=\phi$: continue\\
 7. \quad \textbf{Else:} \textbf{For} each segmentation $s_j$ in $N_i$:\\
 8. \qquad 	Finding its spatiotemporal neighbor set that satisfy \\ \qquad\quad both spatial adjacency and similar temporal \\ \qquad\quad patterns $(r>\epsilon):N_j$\\
 9. \qquad $R_j=[\quad ]$\\
 10. \qquad \textbf{For} each segmentation $s_k$ in $N_j$:\\
 11. \quad\qquad Append $r_{kj}$ to $R_j$\\
 12. \qquad \textbf{If} $max(R_j)>r_{ij}$:\\
 13. \quad\qquad remove $s_j$ from $N_i$\\
 14. \quad \textbf{If} $N_i\neq\phi$: Label each segmentation in $N_i$ as 1\\ 
 15. \quad \textbf{Else:} $l_i=1$\\
 16. \quad Getting the aggregated region $C_i=[x_i,N_i]$\\
 17. Getting the aggregated region set:\\ \quad\quad\quad $S_{agg}=[C_1,C_2,\cdots,C_d ](d<m)$\\
 \hline
\end{tabular}
\end{table}

The core component of graph generation method is aggregating regions step. In simple terms, this step like the process of mutual attraction between particles. Aggregated regions are similar to particles while correlation coefficients between different regions are similar to attractive forces between particles. The aggregation between the particles is within a certain spatial range, the more attractive together, just like the aggregating regions step in our proposed graph generation method.
\begin{figure}[h]
  \centering
  \includegraphics[width=1\linewidth]{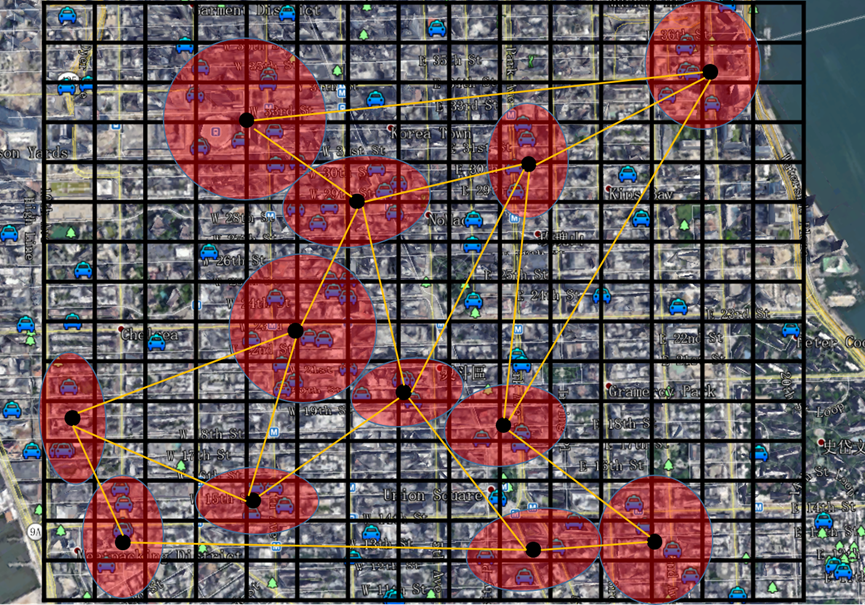}
  \setlength{\abovecaptionskip}{-10pt}
  \setlength{\belowcaptionskip}{-10pt}
  \caption{Some regions are aggregated according to spatial proximity and temporal pattern similarity. After aggregating step, we can model different modes of graph between aggregated regions.}
\end{figure}

\subsection{Short-term temporal dynamics view}
Although RNN structures are widely used in time series data processing, it still suffers from time consumption and gradient optimization. To capture short-term temporal dynamics, 1D CNN model can be taken into consideration and it can achieve comparable performance with lower computation burden in many scenes. The short-term temporal dynamics view contains two 1-D CNN units followed by gated linear units (GLU) as a non-linearity activation function. For each virtual node, 1D CNN explores \emph{k} temporal neighbors of input time series with padding to keep the length of sequences. The input of short-term temporal dynamics view is $X_1\in R^{M\times N\times C_1}$. N represents the number of virtual nodes. Hence, the input of 1D CNN for each virtual node is a length-M sequence with $C_1$ channels as $x_1^i\in R^{M\times C_1}$. Note that, to enhance the representation of the input sequence, we expand the dimension of the original input $X_0\in R^{M\times N\times C_0} (C_0=1)$ before 1D CNN operation by a embedding layer. The embedding layer is actually a fully connected feedforward layer $w_0\in R^{C_0\times C_1}$.
\begin{equation}
X_0 w_0=X_1\in R^{M\times N\times C_1}
\end{equation}
The convolution kernels $\Gamma_1,\Gamma_2\in R^{K\times C_1\times C_2 }$ are designed to map the feature vector of each virtual node $x_1^i$ to P and Q $(P,Q\in R^{M\times C_2})$. The gated 1D CNN can be defined as:
\begin{equation}
\begin{aligned}
X_2&=\Vert_{i=1}^{N} P\odot \sigma(Q)\\
&=\Vert_{i=1}^{N}[(\Gamma_1*x_1^i )\odot \sigma(\Gamma_2*x_1^i )]\in R^{M\times N\times C_2}
\end{aligned}
\end{equation}
Where P, Q are input of gates in GLU respectively, notation $\odot$ represents element-wise Hadamard product and $\Vert$ represents concentration operation. The sigmoid gate $\sigma(Q)$ controls retention ratio of information P. In this case, we apply the same convolution kernels $\Gamma_1$  and $\Gamma_2$ to each virtual node equally. The structure of short-term temporal dynamics view is displayed in Fig 3.
\begin{figure}[h]
  \centering
  \includegraphics[width=1\linewidth]{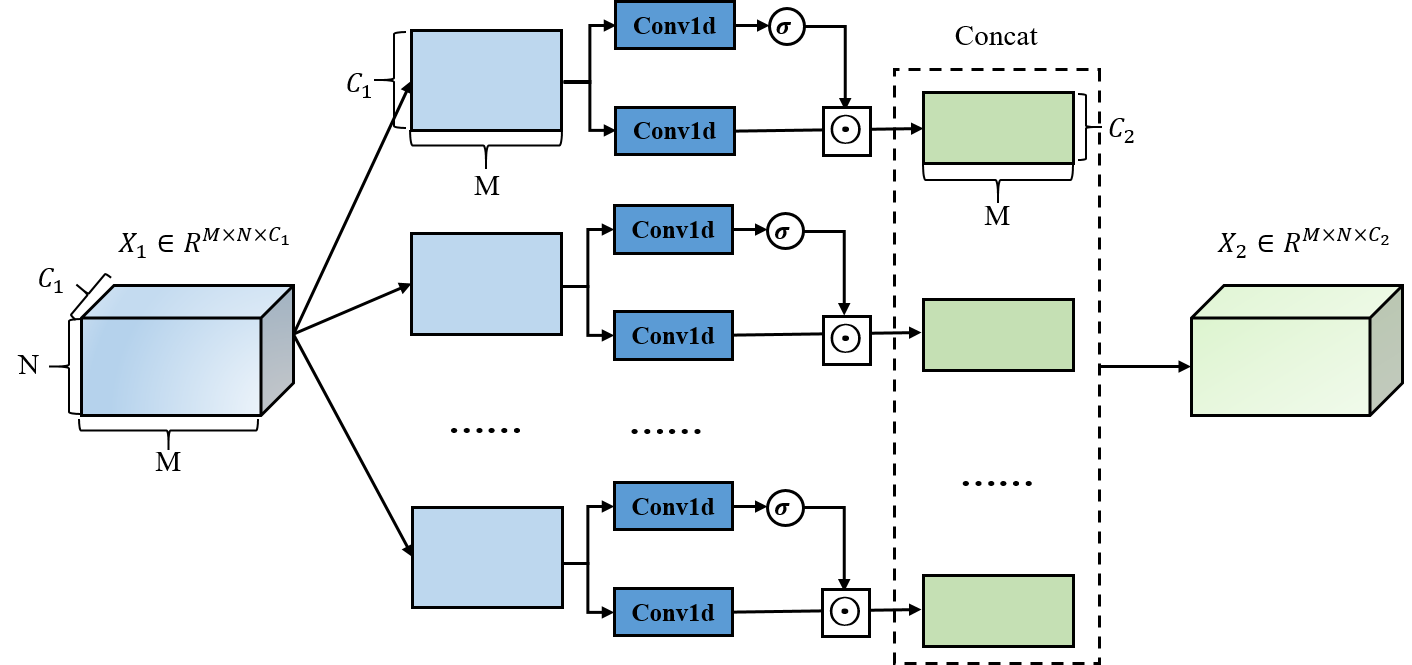}
  \setlength{\abovecaptionskip}{-10pt}
  \setlength{\belowcaptionskip}{0pt}
  \caption{The overview of the short-term temporal dynamics view.}
\end{figure}

\subsection{Spatial dynamics view}
In most previous works, ChebNet and Diffusion-Net are two common GCN model to capture spatial dynamics on traffic network. To simplify computation and consider the spatial physical meaning, we utilize spatial GCN model in this case. GAT is a fruitful spatial GCN model to capture spatial dependences, which adaptively calculates contribution of neighbor regions by attention mechanism. In this case, we generalize the traditional 2-D GAT to 3-D GAT, which has an additional time step length dimension M. Where N represents the number of nodes, $C_2$ represents the dimension of feature map calculated by short-term temporal dynamics view, $C_3$ represents the dimension of output feature map by 3-D GAT.

\noindent In GAT model, the attention factor is computed as Softmax form:
\begin{equation}
a_{ij}=\frac{exp(LeakyReLu(\varphi^T [\omega h_i\Vert\omega h_j]))}{\sum_{j\in N(i)}exp(LeakyReLu(\varphi^T [\omega h_i\Vert\omega h_j]))}
\end{equation}
Where $\varphi\in R^{M\times2C_3\times1}$ and $\omega\in R^{M\times C_2\times C_3}$ are learnable parameters, $h_i\in R^{M\times1\times C_2}$ represents the input latent representation of node i.
To stabilize the training process, we can also involve multi-head attention mechanism. The multi-head attention mechanism here is to accumulate $L_G$ individual GAT output results:
\begin{equation}
X_{out}^g=\Vert_{i=1}^{N} [LeakyReLu(\frac{1}{L_G}\sum_{l=1}^{L_G}\sum_{j\in N(i)}\alpha_{ij}^k\omega^k h_j)]
\end{equation}
For multi-graph, we use different two-layer multi-head GAT models to capture their spatial dependences and concentrate their latent representation together:
\begin{equation}
X_3=\Vert(X_{out}^1,\cdots,X_{out}^g)\in R^{M\times N\times g\centerdot C_3}
\end{equation}
In addition, to fully fuse high-level information and low-level information, from figure 1, we design a residual connection between the input of short-term temporal dynamics view and the output of spatial dynamics view. The residual connection is to sum high-level information and low-level information directly:
\begin{equation}
X_4=(X_1w_r+X_3)\in R^{M\times N\times g\centerdot C_3}
\end{equation}
Note that, $w_r\in R^{C_1\times g\centerdot C_3}$ is a fully connected operation for dimensional consistency. The notation g is the number of graph types. If we can model these three kinds of graphs from our datasets, g equals 3 in this case.
\subsection{Long-term temporal dynamics view}
Long-term temporal dynamics view is actually a Transformer model, which is composed of self-attention layer, position encoding mechanism and feedforward output layer. Similar to the 1D CNN in short-term temporal dynamics view, the transformer layer is also applied to each node individually.

To enhance the recognition of sequence position information, position encoding mechanism is involved in Transformer layer. The position encoding mechanism is to conduct sine encoding on even positions as well as cosine encoding on odd positions. The position encoding mechanism is defined as:
\begin{equation}
\begin{aligned}
&d_{model}=g\centerdot C_3\\
&pe_m^i=\left\{\begin{array}{ll}
sin(m/(10000^{2d/d_{model}})) &\text {if }m=0,2,\cdots \\
cos(m/(10000^{2d/d_{model}})) &\text {if }m=1,3,\cdots
\end{array}\right.\\
&PE_i=(\Vert_{m=0}^{M} (\Vert_{d=0}^{d_{model}} pe_m^i)) \in R^{M\times gC_3}
\end{aligned}
\end{equation}
In the case of involving position encoding, the input of the self-attention layer needs to add the position encoding variable on the basis of the original input, which is defined as:
\begin{equation}
\begin{aligned}
\tilde{x}_4^i&=x_4^i+PE_i,\quad\tilde{X}_4&=\Vert_{i=1}^{N} \tilde{x}_4^i
\end{aligned}
\end{equation}
For single head self-attention layer, there are commonly three types of vectors, queries, keys and values for all the nodes, corresponding to the notation $Q,K,V\in R^{M\times C_1}$, which is defined as:
\begin{equation}
SA(Q,K,V)=softmax(\frac{QK^T}{\sqrt{d_k}})V
\end{equation}
$d_k$ is a normalization factor and its value is consistent with the feature dimension of Q. The total input of self-attention layer is $\tilde{X}_4$ from spatial dynamics view but considering that self-attention layer is for an independent node, we should divide $\tilde{X}_4\in R^{M\times N\times gC_3}$ into N $\tilde{x}_4^i\in R^{M\times gC_3}$ as input for each node. The three types of vectors in the equation (11) are obtained by multiplying the input $x_3$ by three different projection matrices, which is defined as:
\begin{equation}
Q=\tilde{x}_4^iw^q, K=\tilde{x}_4^iw^k, V=\tilde{x}_4^iw^v
\end{equation}
Where $w^q,w^k,w^v\in R^{2C_3\times d_k}$ are there types of learnable projection matrix for Q, K and V. To enhance capability of the model, we adopt multi-head self-attention (MHSA) in this case, which is defined as:
\begin{equation}
\begin{aligned}
&mh_i=MHSA(\tilde{x}_4^i)=(\Vert_{l=1}^{L_T} (softmax(\frac{(\tilde{x}_4^iw_j^q)(\tilde{x}_4^iw_j^k)^T}{\sqrt{d_k}})\\
&(\tilde{x}_4^iw_j^v)))\centerdot w^p,\quad X_5=(\Vert_{i=1}^{N} mh_i) \in R^{M\times N\times d_p}
\end{aligned}
\end{equation}
Different from multi-head GAT, we concentrate $L_T$ outputs of single self-attention in this case. The notation $w_j^q,w_j^k,w_j^v$ are three types of projection matrix in the $j_{th}$ attention head. And $w^p\in R^{M\times d_p}$ is another projection matrix for output of multi-head self-attention.

Following the self-attention layer is a two-layer feedforward neural network, which is defined as:
\begin{equation}
\hat{Y}_{t+T}=ReLU(ReLU(BN(X_5 w_{f1} )) w_{f2} )
\end{equation}
Where $\hat{Y}_{t+T}$ is the final output of long-term temporal dynamics view, BN is batch normalization operation, $w_{f1}\in R^{d_p\times d_{f1}}$ is the weight of the first FNN layer, $w_{f2}\in R^{d_{f1}\times1}$ is the weight the of the second FNN layer. The overview of long-term temporal dynamics view is displayed in Fig 4.

\begin{figure}[h]
  \centering
  \includegraphics[width=1\linewidth]{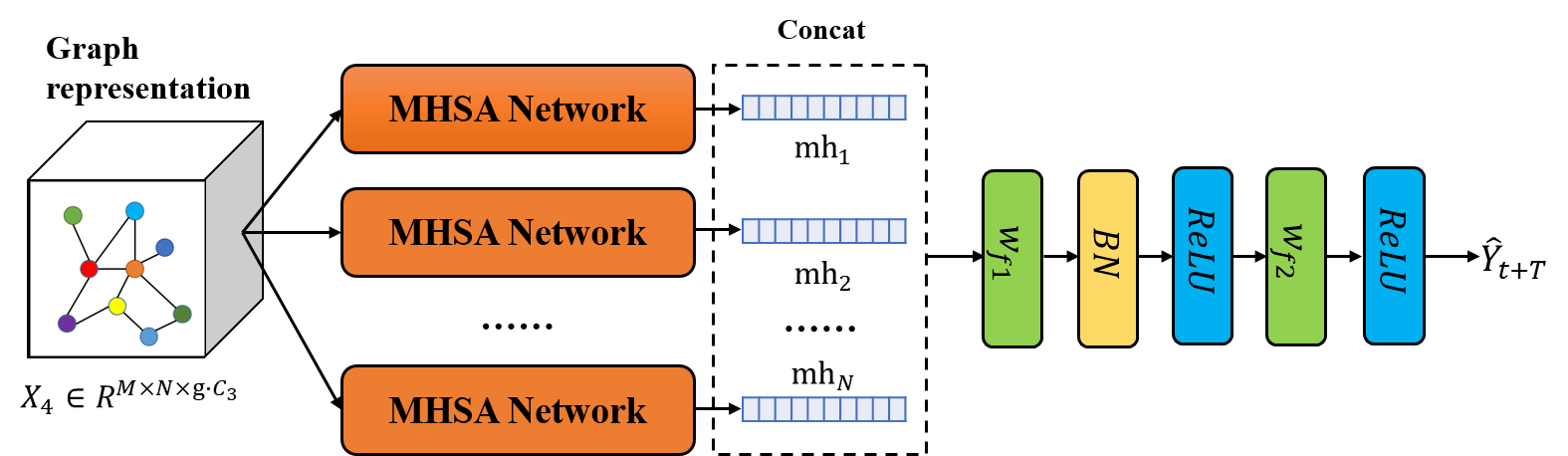}
  \setlength{\abovecaptionskip}{-10pt}
  \setlength{\belowcaptionskip}{0pt}
  \caption{The overview of the long-term temporal dynamics view.}
\end{figure}

To stabilize the gradient descent process of model training, we apply Smooth L1 loss in this case, which is defined as:
\begin{equation}
\begin{aligned}
L&=Smooth L_1 (Y_{t+T}-\hat{Y}_{t+T})\\
Smooth L_1&=\left\{\begin{array}{ll}
0.5y^2 & \text { if } |y|<1 \\
|y|-0.5 & \text { otherwise }
\end{array}\right.
\end{aligned}
\end{equation}

\section{Experiments and Analysis}
\subsection{Dataset Description}
We evaluate our model on two real-world ride-hailing demand datasets, New York City Uber Data and New York City Taxi Data. These two Dataset and New York City Taxi Dataset were collected from whole New York City area. To avoid some abnormal points, we remove some urban fringe areas to limit the geographic range to the latitude interval [40.628, 40.830] and longitude interval [-74.05, -73.88]. The ride-hailing demand data are aggregated per hour as a time slot. The time period of Uber is from 1st April to 31st August, 2014. The time period of taxi is from 1st July to 31st November, 2013. The datasets are split in chronological order with $80\%$ for training, $10\%$ for validation and $10\%$ for testing.

According to Graph generation method in Table 1, we first divided the New York City area by 40*30 grid, which is higher granularity compared with some previous works. In the previous work \cite{saxena2019d-gan,yao2019learning,lin2019spatial-temporal}, New York City area is only divided into 9*9, 16*12 and 20*10 grid map respectively. Then we discard some sparse areas with a certain threshold $\delta$ and aggregate the neighbor regions by Pearson similarity threshold $\epsilon$ to construct multi-graph.
\begin{table}[ht]
\setlength{\abovecaptionskip}{-0.1cm}
\centering
\caption{The parameters for virtual graph construction}
\label{tab:dataset}
\begin{tabular}{ccccc}
\hline
Dataset &  $\delta$ &  $\epsilon$ &  Number of &  Number of\\
&          &          & significant regions & virtual nodes\\
\hline
NYC Uber      & 1   & 0.5 & 148 & 102\\
NYC Taxi   & 1   & 0.5  & 315 & 225\\
\hline
\end{tabular}
\end{table}

For NYC Taxi dataset, geographic information, corresponding timestamps and OD information are all available. Hence, we can model the three different graphs: distance graph, correlation graph and mobility graph. But for NYC Uber dataset, the OD information is unavailable, so we only construct distance graph and correlation graph in this case.

\subsection{Training Details and Baselines}
We ran our model on Nvidia Tesla P40 GPU. The Adam optimizer is used in training process of DMVST-GNN. The initial learning rate with the epoch-by-epoch decay is set as 0.005 and the decay rate is set as 0.95. The length of input sequence is set as 12. The number of attention head in GAT and Transformer are both set as 4. The hidden dimension of embedding layer is as 48. The hidden dimension of gate 1D-CNN is set as 48. The hidden dimensions of two-layer GATs are both set as 24. The hidden dimensions of self-attention layer and FNN in Transformer model are as 72 and 256 respectively. The batch is set as 16 in all subsequent experiments. And we use early stop strategy in training process to prevent over-fitting.

In comparative experiment step, nine baselines are presented to compare with DMVST-GNN. Note that, our predict target is significant ride-hailing demand region in this paper, so we only need to evaluate experiment results in these retained and aggregated regions. The baseline models are shown below:

\noindent \textbf{ARIMA} is the classical time series mathematical modeling model \cite{han2004real}.

\noindent \textbf{Random Forest} is a ensemble statistical learning model \cite{leshem2007traffic}.

\noindent \textbf{Conv-LSTM} is a coupled model of CNN and LSTM, which conduct gate operation by convolutional layer \cite{xingjian2015convolutional}.

\noindent \textbf{ST-ResNet} is a CNN-based model, which capture the trend, the periodicity and the closeness dynamics by ResNet and fuse them together \cite{zhang2017deep}.

\noindent \textbf{DMVST-Net} is a hybrid model that combines spatial view, temporal view, and semantic view, which integrates CNN, LSTM and graph embedding \cite{yao2018deep}.

\noindent \textbf{DCRNN} is a coupled model of Diffusion GCN and GRUN, which is the first state-of-art distance graph based model applied in road sensor network \cite{li2017diffusion}. In this paper, DCRNN is deployed on distance graph.

\noindent \textbf{Graph WaveNet} is a hybrid model that combines Diffusion GCN and 1D Gate CNN, capturing spatial dependences by Diffusion GCN and temporal dependences by 1D Gate CNN \cite{yu2018spatio-temporal}. Graph WaveNet is also designed based on distance graph.

\noindent \textbf{Multi-GCN} is a hybrid model of GCN and LSTM-based Seq2seq model that first involves multi-type spatial graph representation by distance, correlation and interaction \cite{chai2018bike}.

\noindent \textbf{ST-MGCN} is also a multi-graph fusion spatiotemporal prediction model, but ST-MGCN involve attention mechanism in contextual gated recurrent neural network to capture temporal dynamics \cite{geng2019spatiotemporal}. Different from the original version, we construct the graphs as distance graph, correlation graph and mobility graph in this case.

\subsection{Comparison for Global Prediction}
We evaluate the effectiveness of proposed framework DMVST-GNN in this sub-section by comparative experiment. Two metrics are used in evaluation the prediction performance of each algorithm, respectively as the Root Mean Square Error (RMSE), Mean Absolute Error (MAE). The global prediction results of comparative experiments for NYC Uber and NYC Taxi are shown in Table 3. We conduct ten dependence experiments to obtain the mean of these two metrics. From table 3, we find that graph-based models achieve superior performance than mathematical model, statistic learning models and pixel-level deep learning models. And our proposed model DMVST-VGNN achieve state-of-art performance among these models. For NYC Uber dataset, DMVST-VGNN low RMSE by $7.14\%$ and low MAE by $7.61\%$ compared with the optimal baseline model. For NYC Taxi dataset, DMVST-VGNN low RMSE by $4.71\%$ and low MAPE by $4.61\%$ compared with the optimal baseline model.

\begin{table}[htpb!]
\setlength{\abovecaptionskip}{-0.1cm}
\caption{Performance comparison of different approaches for NYC Uber and NYC Taxi significant ride-hailing demand prediction.}
\centering
\begin{tabular}{|l|l|l|l|l|}
\hline
\multirow{2}{*}{Algorithms} & \multicolumn{2}{l|}{NYC Uber} & \multicolumn{2}{l|}{NYC Taxi} \\ \cline{2-5}
       & RMSE           & MAE          & RMSE           & MAE          \\ \hline
ARIMA & 4.48 & 1.92 & 34.68 & 16.01 \\ \hline
Random Forest & 3.97 & 1.74 & 31.74  & 14.27 \\ \hline
Conv-LSTM & 3.11 & 1.43 & 29.56 & 12.69\\ \hline
ST-ResNet & 2.89 & 1.32 & 27.44 & 11.37 \\ \hline
DMVST-Net & 2.45 & 1.13 & 26.76 & 10.99 \\ \hline
DCRNN & 2.51 & 1.17 & 27.11 & 11.23 \\ \hline
Graph WaveNet  & 2.32 & 1.08 &26.23  & 10.81 \\ \hline
Multi-GCN & 2.26 & 1.06 &26.12  & 10.77 \\ \hline
ST-MGCN & 1.96 & 0.92 &24.22  & 9.98 \\ \hline
DMVST-VGNN & \textbf{1.82} & \textbf{0.85} & \textbf{23.08} & \textbf{9.52}\\ \hline
\end{tabular}
\end{table}

\subsection{Component Analysis}
To further investigate the effect of different modules of DMVST-GNN, we design eight variants of the DMVST-GNN model. We compare these eight variants with the DMVST-GNN model on NYC Taxi dataset to respectively demonstrate effectiveness of spatial dependencies learning and temporal dependencies learning. The details of these eight models are described as below:\\
1.	\textbf{D-GNN:} This model only retains the distance graph in spatial dynamics view. There is no change in short-term dynamics view and long-term dynamics view.\\
2.	\textbf{C-GNN:} This model only retains the correlation graph in spatial dynamics view. There is no change in short-term dynamics view and long-term dynamics view.\\
3.	\textbf{M-GNN:} This model only retains the mobility graph in spatial dynamics view. There is no change in short-term dynamics view and long-term dynamics view.\\
4.	\textbf{D+C-GNN:} This model retains the distance and correlation graph in spatial dynamics view. There is no change in short-term dynamics view and long-term dynamics view.\\
5.	\textbf{C+M-GNN:} This model retains the mobility and correlation graph in spatial dynamics view. There is no change in short-term dynamics view and long-term dynamics view.\\
6.	\textbf{D+M-GNN:} This model retains the distance and mobility graph in spatial dynamics view. There is no change in short-term dynamics view and long-term dynamics view.\\
7.	\textbf{SD-GNN:} This model removes the long-term temporal dynamics view from DMVST-GNN. And there is no change in spatial dynamics view and short-term temporal dynamics view.\\
8.	\textbf{LD-GNN:} This model removes the short-term temporal dynamics view from DMVST-GNN. And there is no change in spatial dynamics view and long-term temporal dynamics view.

Fig 5(a) reveal the effectiveness of multi-graph structure for spatial dependencies learning. Fig 5(b) reveal that the combination of the short-term dynamic module and the long-term dynamic module is effective for temporal dependencies learning.
\begin{figure}[h]
  \centering
  \includegraphics[width=1\linewidth]{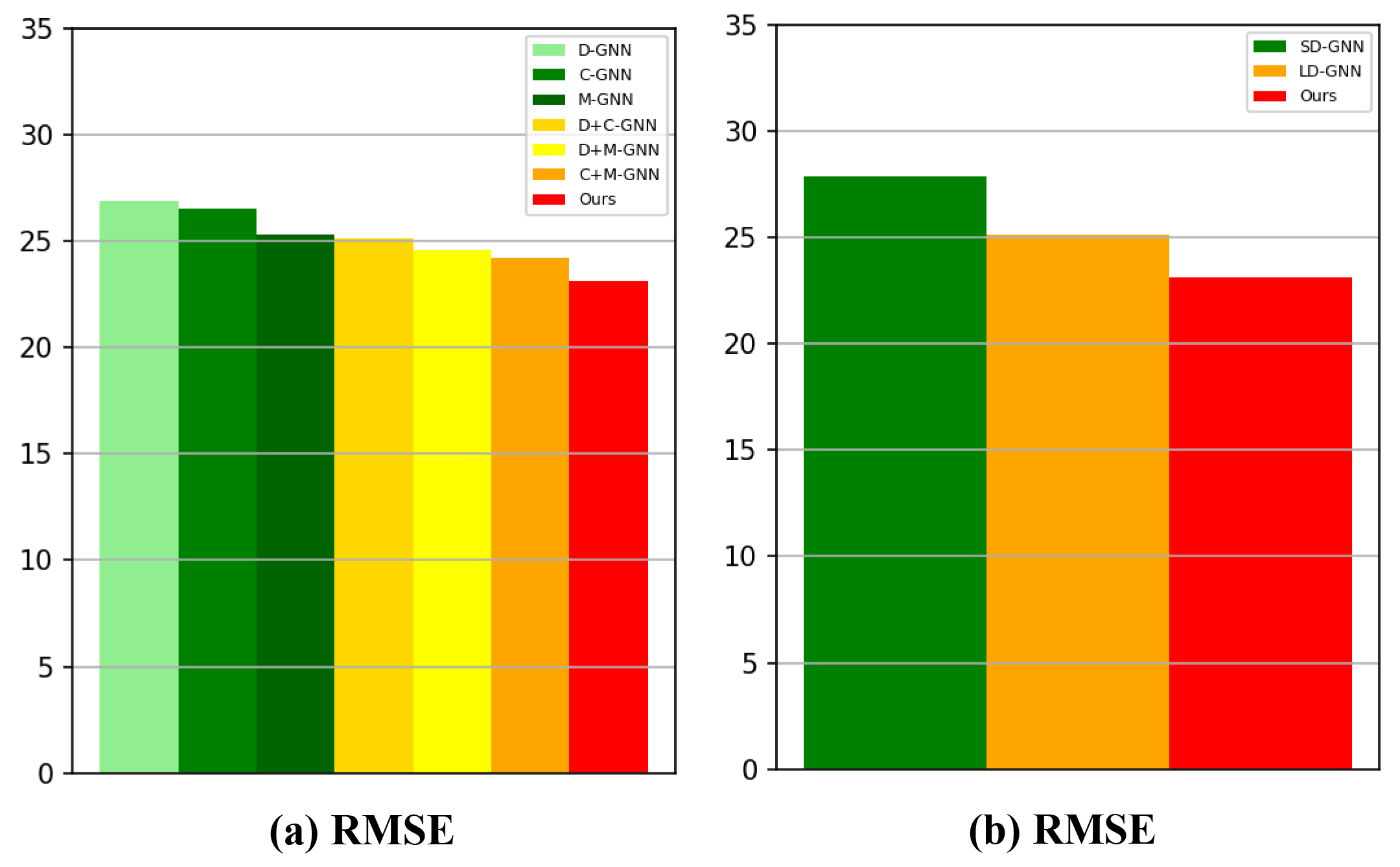}
  \setlength{\abovecaptionskip}{-10pt}
  \setlength{\belowcaptionskip}{0pt}
  \caption{The comparative experimental results of RMSE for DMVST-GNN and eight variants.}
\end{figure}


\subsection{Top K Hot Regions Prediction}
Usually, the prediction accuracy of the hot demand regions is our most concerned about. To verify effectiveness of our proposed model in some hot demand regions, we compare the baseline models with DMVST-VGNN for top $10\%$ hot demand regions on NYC Uber and NYC Taxi datasets respectively. In this case, we select Mean Absolute Percent Error (MAPE) to measure the mean error rate in top K regions. From Table 4, we find that DMVST-VGNN model respectively low MAPE by $3.51\%$ and $5.99\%$ on two datasets compared with the optimal baseline ST-MGCN. This means that our model is superior to other baselines for both the global prediction and hot demand regions prediction.
\begin{table}[htpb!]
\setlength{\abovecaptionskip}{-0.1cm}
\caption{Performance comparison of top $10\%$ hot regions for NYC Uber and NYC Taxi significant ride-hailing demand prediction.}
\centering
\begin{tabular}{|l|l|l|}
\hline
Algorithms & MAPE(Uber) & MAPE(Taxi) \\ \hline
ARIMA & 0.1561  &  0.1710  \\ \hline
Random Forest & 0.1482  & 0.1645   \\ \hline
Conv-LSTM & 0.1374  & 0.1511  \\ \hline
ST-ResNet & 0.1315  & 0.1436   \\ \hline
DMVST-Net & 0.1281  & 0.1401  \\ \hline
DCRNN & 0.1292 & 0.1405 \\ \hline
Graph WaveNet  & 0.1271 & 0.1392 \\ \hline
Multi-GCN & 0.1265 & 0.1385 \\ \hline
ST-MGCN & 0.1196 & 0.1301 \\ \hline
DMVST-VGNN & \textbf{0.1154} &\textbf{0.1223}  \\ \hline
\end{tabular}
\end{table}

\section{Conclusion}
In this paper, we propose DMVST-VGNN, a framework to forecast significant citywide ride-hailing demand, which overcomes the limitations of spatial data sparsity in fine-grained prediction by constructing multiple interpretable virtual graphs. Also, both long-term and short-term temporal dynamics are taken into consideration to enhance the learning capabilities for long sequences. The experimental results demonstrate the superiority of our proposed model compared with some state-of-art baseline models. In the future, we will extend our proposed framework to other spatiotemporal prediction tasks, for instance, air pollution forecasting and epidemic forecasting.
\newpage
\bibliographystyle{aaai}
\bibliography{myref}

\end{document}